\pdfoutput=1

\documentclass[11pt]{article}

\usepackage[preprint]{acl}

\usepackage{times}
\usepackage{latexsym}
\usepackage{multirow}
\usepackage[T1]{fontenc}

\usepackage[utf8]{inputenc}

\usepackage{microtype}

\usepackage{inconsolata}
\usepackage{pifont}
\usepackage{graphicx}
\usepackage{amssymb} 
\usepackage{float}
\usepackage{array}
\usepackage{booktabs}
\usepackage{enumitem}
\usepackage{amsmath}
\usepackage{algorithm}
\usepackage{algpseudocode}
\usepackage{subcaption}

\title{AppAgentX: Evolving GUI Agents as Proficient Smartphone Users}

\author{Wenjia Jiang$^{1,2}$, 
        Yangyang Zhuang$^{1}$, 
        Chenxi Song$^{1}$,
        Xu Yang$^{3}$,
        Joey Tianyi Zhou$^{4,5}$
        Chi Zhang$^{1}$, 
         \\ 
        $^{1}$Westlake University, China,
        $^{2}$Henan University, China, 
        $^{3}$Southeast University, China \\
        $^{4}$IHPC, Agency for Science, Technology and Research, Singapore \\
        $^{5}$CFAR, Agency for Science, Technology and Research, Singapore\\
        \texttt{\{jiangwenjia, chizhang\}@westlake.edu.cn}, \\
        \url{https://appagentx.github.io/}
    }

\begin{document}

\twocolumn[{
\renewcommand\twocolumn[1][]{#1}
\maketitle
\centering
\vspace{-1.5cm}
\includegraphics[width=\linewidth]{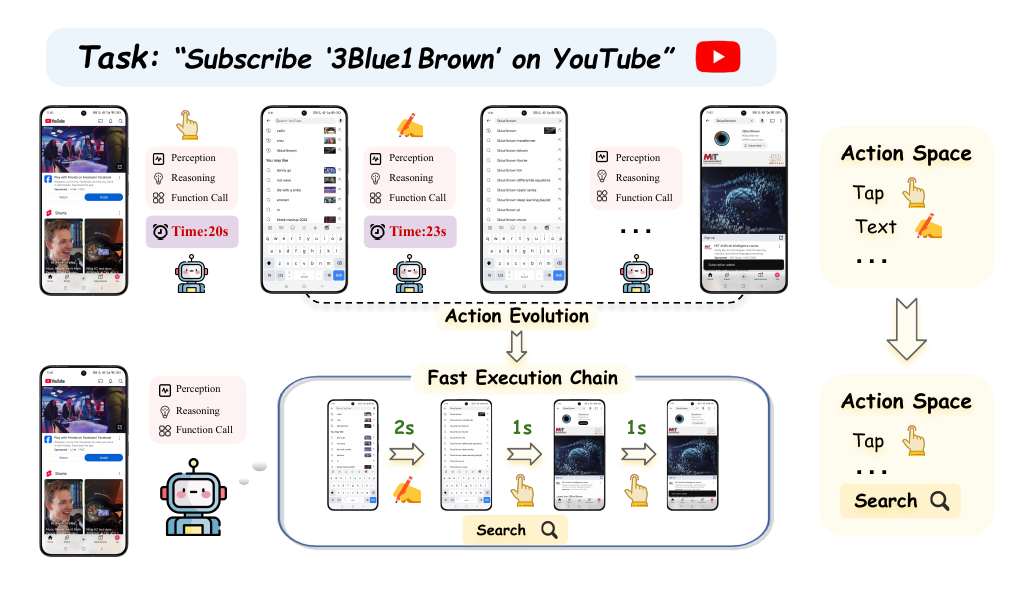}
\captionsetup{type=figure}
\vspace{-2.5em}
\caption{\textbf{Illustration of the proposed evolutionary mechanism in GUI agents.} The agent evolves a high-level action, "Search", which replaces a sequence of inefficient low-level actions. This evolution eliminates the need for step-by-step reasoning, significantly enhancing the agent's efficiency.}

\vspace{0.3cm}
\label{fig:motivation_example}
}]


\begin{abstract}
Recent advancements in Large Language Models (LLMs) have led to the development of intelligent LLM-based agents capable of interacting with graphical user interfaces (GUIs). These agents demonstrate strong reasoning and adaptability, enabling them to perform complex tasks that traditionally required predefined rules. However, the reliance on step-by-step reasoning in LLM-based agents often results in inefficiencies, particularly for routine tasks. In contrast, traditional rule-based systems excel in efficiency but lack the intelligence and flexibility to adapt to novel scenarios.
To address this challenge, we propose a novel evolutionary framework for GUI agents that enhances operational efficiency while retaining intelligence and flexibility. Our approach incorporates a memory mechanism that records the agent's task execution history. By analyzing this history, the agent identifies repetitive action sequences and evolves high-level actions that act as shortcuts, replacing these low-level operations and improving efficiency. This allows the agent to focus on tasks requiring more complex reasoning, while simplifying routine actions.
Experimental results on multiple benchmark tasks demonstrate that our approach significantly outperforms existing methods in both efficiency and accuracy. The code will be open-sourced to support further research.
\end{abstract}
\section{Introduction}
Recent advancements in artificial intelligence have been significantly influenced by the development of LLMs such as GPT-4 \cite{GPT-4} and DeepSeek-V3 \cite{DeepSeek-V3}. These models, renowned for their ability to process and generate human-like text, have catalyzed innovations across various domains, including natural language understanding, generation, and reasoning. One promising application is the creation of LLM-based agents, which use natural language to autonomously interact with users and perform a wide range of tasks. Unlike traditional rule-based systems, LLM-based agents exhibit the flexibility to understand complex tasks and generalize to novel scenarios, enhancing human-computer interaction.

A notable development in this area is the emergence of LLM-based agents capable of operating graphical user interfaces (GUIs). Unlike Robotic Process Automation (RPA), which relies on predefined rules, these agents can engage with GUIs dynamically, mimicking human-like interactions by outputting low-level actions such as clicks, drags, and scrolls. This approach not only increases operational flexibility but also enables the execution of tasks that require adaptability and reasoning, without needing access to backend systems or APIs.

To achieve optimal performance, various mechanisms have been introduced to guide the LLMs in generating accurate and contextually appropriate actions. 
Techniques such as reflection\cite{reflection} and chain-of-thought\cite{cot} reasoning have been employed to help the model carefully consider the implications of each action before execution.
While these methods can significantly enhance the agent’s decision-making abilities, they often come at the cost of efficiency, particularly for tasks that do not require advanced reasoning. 
For example, as illustrated in Figure~\ref{fig:motivation_example}, during simple tasks like search operations, the agent may need to reason each step, such as clicking the search box, typing text, and pressing submit, leading to unnecessary delays. While traditional RPA methods can execute these fixed steps rapidly, they lack the flexibility to handle tasks requiring intelligent judgment. This highlights the need for a more efficient, adaptable approach to automate routine tasks.

In response to this challenge, we propose a novel, evolving framework for GUI agents that aims to enhance both the efficiency and intelligence of the agent’s behavior. 

Our approach enables the agent to learn from their past interactions and dynamically evolve more abstract, high-level actions, eliminating the need for repetitive low-level operations.
Specifically, the agent will analyze its execution history to identify patterns in repetitive, low-intelligence actions, such as those involved in routine tasks. 
From this analysis, the agent can generate a high-level action, encapsulating a series of low-level actions, enabling it to perform tasks more efficiently. 
For example, in a search operation, the agent would automatically evolve a "search" action that directly executes the required sequence, improving both speed and accuracy. 

Intuitively, our framework enables the agent to focus its reasoning capabilities on tasks requiring more intelligent judgments, while simplifying repetitive tasks into a more compact form. 
To support this approach, we design a knowledge base structured as a chain to record task execution history and facilitate the abstraction and evolution of behaviors. This knowledge base allows the agent to continuously improve its task execution strategies, further optimizing its performance and enabling the evolution of more intelligent, high-level actions over time.

Our approach relies entirely on visual information, eliminating the need for backend access or APIs. Extensive experiments show that it outperforms baseline and state-of-the-art (SOTA) methods in both efficiency and accuracy across several benchmark tasks. 

The main contributions of this paper are summarized as follows:
\begin{itemize}[itemsep=1pt, topsep=2pt]
    \item We propose an evolutionary mechanism that enables a GUI Agent to learn from its task execution history and improve its efficiency by abstracting repetitive operations.
    \item We design a chain-based framework for recording and optimizing the agent's execution behavior.
    \item Our code will be open-sourced to facilitate further research in this area.
\end{itemize}
\section{Related Works}

\noindent\textbf{Large language models.}
Recent advancements in large language models (LLMs) have significantly expanded the scope of AI-driven automation. Models such as GPT-4 \cite{GPT-4} and DeepSeek-V3 \cite{DeepSeek-V3} exhibit strong natural language understanding and reasoning abilities. These capabilities allow LLMs to process complex UI structures and facilitate interactive decision-making, forming the foundation for LLM-driven GUI Agents \cite{a-comprehensive-overview}. Unlike traditional script-based or rule-based approaches \cite{Script-Based-Automation, Rule-based-exploratory}, LLM-powered agents can generalize across diverse applications and dynamic interfaces without explicitly predefined rules. However, challenges remain in model efficiency, adaptability, and spatial reasoning, necessitating further optimization in both architectural design and training methodologies.

\noindent\textbf{LLMs as Agents.}
LLMs have significantly advanced intelligent agents, enabling complex task execution. AutoGPT\cite{AutoGPT}, AFlow\cite{AFlow}, MetaGPT\cite{MetaGPT}, and AutoAgent\cite{AutoAgents} exemplify autonomous task decomposition and execution, while Stanford Smallville\cite{Generative-Agents} and Agent Hospital\cite{Agent-Hospital} showcase multi-agent simulations. LLM-driven multimodal agents also enhance perception and decision-making across domains. GPT-Driver\cite{GPT-Driver} enables adaptive motion planning for autonomous driving, SmartPlay\cite{SmartPlay} improves agent intelligence in gaming, and MP5\cite{MP5} integrates active perception for efficient task execution in robotics.

\noindent\textbf{LLMs as GUI Agents.}
Beyond general agents, LLMs have also enhanced GUI automation, surpassing traditional script-based methods in flexibility and adaptability. WebVoyager\cite{WebVoyager}, AppAgent\cite{AppAgent}, and MobileAgent\cite{MobileAgent} leverage multimodal perception for interactive interfaces, while UFO\cite{UFO}, AutoGLM\cite{AutoGLM}, and MMAC-Copilot\cite{MMAC-Copilot} improve cross-platform adaptability and multi-agent collaboration. To refine UI understanding, OmniParser\cite{OmniParser} and Ferret-UI\cite{Ferret-UI} enhance element recognition, while TinyClick\cite{TinyClick} and CogAgent\cite{CogAgent} improve interaction precision and vision-based task execution. Additionally, WebGUM\cite{WebGUM}, MobileVLMS\cite{MobileVLMS}, and DigiRL\cite{DigiRL} optimize performance in dynamic web and mobile environments. 
However, most existing agents rely on static training data and lack continuous adaptation. To address this limitation, we propose AppAgentX, integrating task trajectory memory to enhance efficiency and adaptability in GUI automation.

\section{Preliminary}
\label{sec:Preliminary}

Before delving into our proposed methodology, we first introduce a baseline method for LLM-based GUI agents. This baseline serves as a foundation for understanding the core components and tasks involved in enabling LLMs to control smartphones.

The process of utilizing LLMs for smartphone control involves two key stages: screen perception and action execution. The screen perception phase begins with capturing a screenshot of the device’s current interface. In order to accurately interpret this screenshot, we employ OmniParser \cite{OmniParser} to detect and label all interactive elements within the interface, such as buttons and text boxes. OmniParser annotates these elements with tagged bounding boxes, which are subsequently overlaid onto the original screenshot for clear visualization.
Following this, the annotated screenshot is passed to the LLM for action planning. At this stage, the LLM interprets the UI components and generates corresponding actions based on its understanding of the interface.

In the second stage, action execution, we follow  AppAgent~\cite{AppAgent} to define a set of low-level actions that the agent can perform within the smartphone environment. These actions include common gestures such as tapping, long-pressing, swiping, text input, and navigating back. 
These actions collectively define a basic, app-agnostic action space to simulate typical human interactions with a smartphone interface. Formally, the low-level action space is defined as follows:
\begin{equation}
  \label{eq:low_level_action_space}
  \mathcal{A}_{\text{basic}} = \{a_{\text{tap}}, a_{\text{long\_press}}, a_{\text{swipe}}, a_{\text{text}}, a_{\text{back}}\},
\end{equation}
where \( \mathcal{A}_{\text{basic}} \) represents the set of atomic actions available to the agent.

\begin{figure}[ht]
    \centering
    \includegraphics[width=1\linewidth]{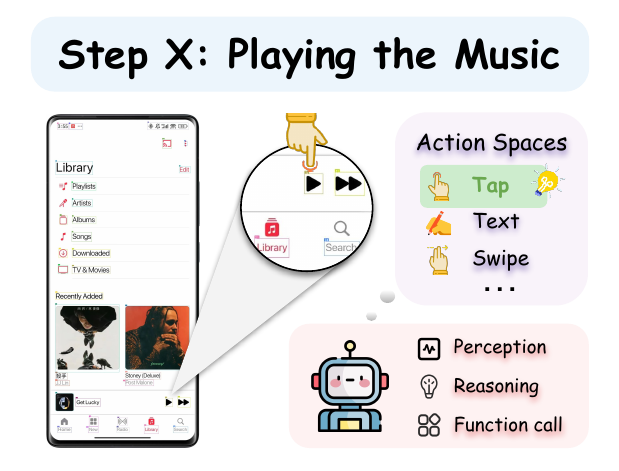}
    \caption{\textbf{Overview of the LLM-based GUI agent baseline.} At each step, the agent captures the current screen of the device and analyzes the interface to select an appropriate action from the predefined action space. The chosen action is then executed to interact with the GUI.}
    \label{fig:OmniBaseline}
\end{figure}

The LLM employs a structured process of observation, reasoning, and function-calling to interact with the smartphone interface. In this process, the LLM iteratively analyzes the current UI, reasons about the appropriate next steps to achieve the desired task, and invokes the corresponding actions from the defined action space. This cycle continues until the task is completed successfully.
An illustration of this process is shown in Figure \ref{fig:OmniBaseline}.

\begin{figure*}[ht]
    \centering
    \includegraphics[width=1\linewidth]{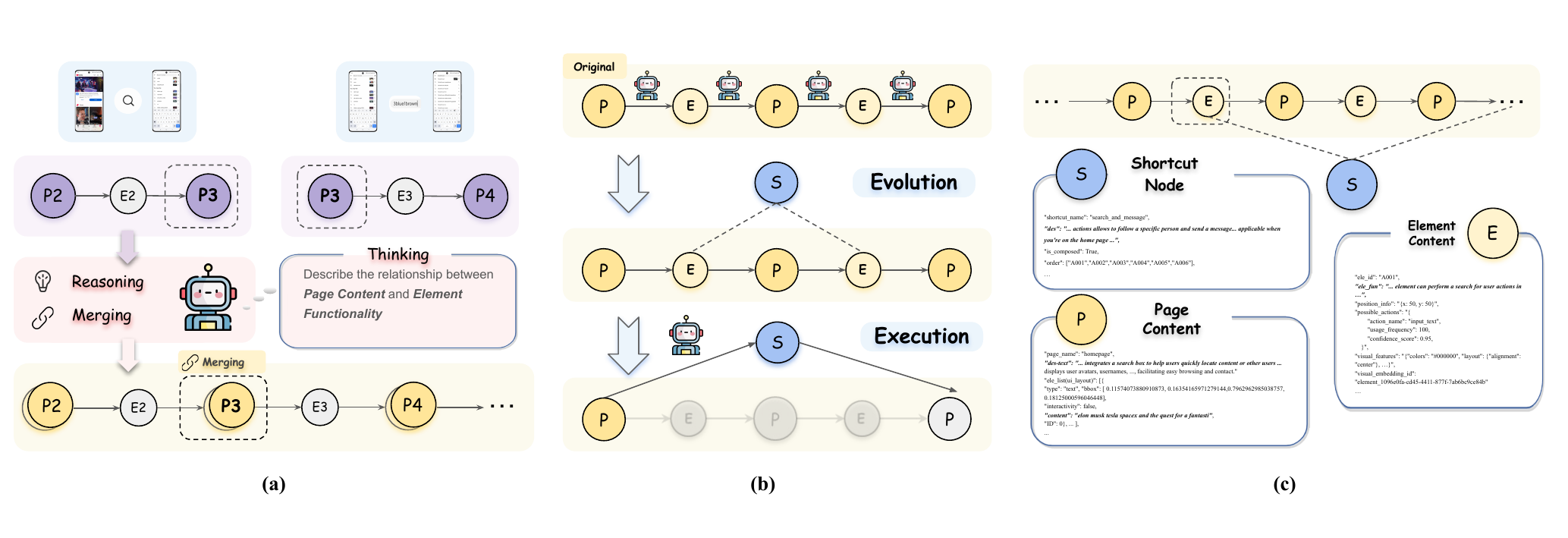}
    \caption{\textbf{Overview of the proposed framework.} (a) The trajectory of a task execution is decomposed into multiple overlapping triples. Based on these triples, the LLM generates functional descriptions of both pages and UI elements. Descriptions of pages that are repeatedly generated are then merged. The entire interaction history is recorded using a chain of nodes. (b) The proposed evolutionary mechanism and execution process. The evolutionary mechanism generates shortcut nodes, which allow the agent to execute a series of actions efficiently without reasoning step by step. (c) An example of the content of nodes within the chain. Each node records essential information, including descriptions of pages, UI elements, and high-level actions, to facilitate understanding of the agent’s interactions.}
    \label{fig:framework}
\end{figure*}

\section{Methodology}
This section outlines the core methodology of the proposed evolutionary framework. The framework comprises three main components: a memory mechanism for recording the agent's operational history, an evolutionary mechanism for improving performance, and an execution strategy for utilizing the evolved agent. We will detail each of these elements in the following subsections.

\subsection{Memory mechanism}
To support the agent's evolution from inefficient to efficient operational modes, it is essential for the agent to retain a record of its past actions and the corresponding outcomes. This enables the agent to learn from its experiences and improve future interactions. To achieve this, we propose a memory mechanism designed to capture and store the agent's trajectory during its interactions with the environment.

The agent's interaction with the UI is modeled as a series of page transitions, where each UI page is represented as a ``page node''. Interactions performed on these pages, such as button clicks or text input, lead to transitions between these nodes. Each page node is characterized by several key attributes, including:

\noindent\ding{226}\textbf{Page Description:} This attribute stores a text description of the entire UI page, initially setting it to empty at the beginning.

\noindent\ding{226}\textbf{Element List:} This property holds a JSON list containing information of all elements detected by OmniParser\cite{OmniParser}, such as the relative screen position, OCR results, \textit{etc.}

\noindent\ding{226}\textbf{Other Properties:} The page nodes also include other properties required for logging, such as page screenshots, ID, timestamps, \textit{etc.}

For more detailed information, we introduce ``element nodes''. Each element node corresponds to a specific UI element, such as a button, text field, or icon. The interactions with these UI elements lead to the page transitions. Similarly, each element node encapsulates essential attributes:

\noindent\ding{226}\textbf{Element Description:} This attribute records a textual description of the element’s functionality, providing a semantic understanding of the element's purpose within the UI. Similar to page descriptions, this attribute is initialized as empty.

\noindent\ding{226}\textbf{Element Visual Embeddings:} This property stores the identifier of the element's screenshot in the vector database. The visual features are extracted using a pre-trained ResNet50 \cite{ResNet50} model.

\noindent\ding{226}\textbf{Interaction Details:}This property includes information related to the basic action for the current element, \textit{e.g.}, tapping, along with the corresponding arguments and other relevant interaction details.

This structure enables the agent to record and learn from both the high-level transitions (from one page to another) and the low-level interactions (with individual UI elements) that lead to those transitions. Figure~\ref{fig:framework} (c) illustrates this process.

\textbf{Extracting Features and Descriptions.}  We next leverage the LLM to generate functional descriptions of pages and individual elements based on observed action sequences. Specifically, we follow the approach in~\cite{AppAgent} by decomposing the agent's trajectory into multiple overlapping triples. Each triple consists of a source page, an action performed on an element, and a target page. These triples capture the changes in page states before and after an action is executed.

The resulting triples are then passed to the LLM for reasoning. The model generates detailed descriptions and functionalities for both the page and element nodes based on the context of the action. This process, illustrated in Figure~\ref{fig:framework} (a), enables the system to build accurate and contextually aware descriptions.

\textbf{Merging Overlapping Descriptions.}
Since the descriptions of page nodes are generated from multiple overlapping action triples, it is likely that the descriptions for the same page node will be generated twice, based on different contexts. Therefore, it is necessary to merge them to generate a unified description for each page node.
To achieve this, we instruct the LLM to combine the descriptions generated from different triples, taking into account both the specific context of each individual action and the broader global task the agent is performing. This allows the LLM to generate a more enriched and complete description of the page, considering its function in the overall task.
The merged descriptions provide a detailed, unified record of the agent’s interactions with the UI, contributing to a coherent chain of node attributes that document the agent’s progress as it completes the task. 

\subsection{Evolutionary Mechanism}
The primary goal of the evolutionary mechanism is to improve the agent's accuracy and efficiency in executing similar tasks by learning from its previous execution histories. 
In typical task execution scenarios, the action sequence may contain repetitive patterns that do not require deep reasoning at each step. 
For example, to search for a content item, the agent may repeatedly perform actions such as tapping the search box, inputting text, and tapping the search button.
By identifying and exploiting these repetitive patterns, the agent can replace inefficient action sequences with higher-level actions that streamline the process, thereby improving overall task efficiency and accuracy.
To achieve this, we introduce the concept of a \textit{shortcut node}, which represents a high-level action that bypasses previously inefficient action sequences.
The shortcut node is designed to  streamline its decision-making process, eliminating unnecessary reasoning process and accelerating task completion. 

To identify whether an action sequence contains repetitive patterns that can be replaced by shortcut nodes, we design mechanisms that evaluate the trajectory of each task. 
Initially, LLM, drawing from prior knowledge, is tasked with determining whether a given task contains repetitive patterns that may be optimized through the introduction of shortcut nodes. 
If the task is deemed to contain such patterns,  the next step involves inspecting the actual trajectory data.
The trajectory data, which includes the descriptions of all relevant pages and elements, is input into the LLM. Upon receiving this data, the model is prompted to generate the description of a shortcut node, specifying the scenarios in which the shortcut node can be invoked and the specific low-level action sequences that it is intended to replace. This process effectively constructs an evolved action space by integrating new, higher-level actions into the agent’s operational repertoire, allowing it to perform tasks more efficiently.

To formally define the construction of high-level actions and the expansion of the action space, we introduce the following representations. A high-level action \( \tilde{a} \) is composed by abstracting a sequence of low-level actions from the basic action set \( \mathcal{A}_{\mathrm{basic}} \).
This abstraction allows the agent to perform more complex tasks with fewer steps, leveraging previously learned patterns of interaction.
We then define the expanded action space \( \mathcal{A}_{\mathrm{evolved}} \), which incorporates both the original low-level actions and the newly introduced high-level actions:
\begin{equation}
  \label{eq:expand_action_space}
  \mathcal{A}_{\mathrm{evolved}} = \mathcal{A}_{\mathrm{basic}} \cup \{\tilde{a}\}.
\end{equation}

Here, \( \mathcal{A}_{\mathrm{evolved}} \) represents the enriched action space that includes both basic, low-level actions and the newly composed high-level actions. This expanded action space enables the agent to perform tasks more efficiently by utilizing higher-level abstractions that reduce the need for repetitive action sequences. 

\subsection{Dynamic Action Execution}
Following the evolution of the action space, the agent can now select high-level actions from the expanded action space to efficiently execute tasks.
\subsubsection{Execution Process}
The high-level execution process begins after the agent captures a screenshot of the current page. At this stage, the system then matches the parsed elements on the page to the stored element nodes in memory, by comparing their visual embeddings. Next, we check whether these identified element nodes are associated with any shortcut nodes. If an association between the element nodes and shortcut nodes exists, the system leverages the LLM to determine whether the corresponding high-level actions can be executed. This decision is made by inspecting the description of the shortcut node in conjunction with the current task context. If the conditions for executing the high-level action are met, the LLM generates an action execution template. This template includes the sequence of low-level actions to be executed by the shortcut node, along with the corresponding function arguments necessary for each action.
In cases where multiple elements on the page are associated with shortcut nodes, the system prioritizes executing the actions based on the order of matching elements. This order is determined by their execution sequence, ensuring that the actions are performed in a logical and efficient manner.
As the sequence actions progresses, partially repetitive operations or even the entire task can be completed more efficiently through the use of high-level actions. 

\subsubsection{Fallback Strategy}
To ensure robustness and task completion reliability, we introduce a fallback strategy that allows the agent to dynamically recover from failures.
If the conditions for executing a high-level action are not met, the agent will default to selecting actions from the basic action space \(\mathcal{A}\). Additionally, in cases where execution errors occur due to incorrect shortcut node matching or unexpected UI responses, the agent reverts to using actions from the basic action space as a fallback. This ensures that the agent can still operate effectively, even when high-level abstractions cannot be applied to the current task.

During the execution of high-level paths, operations that would traditionally require multiple reasoning steps by the LLM are transformed into a page-matching and retrieval-based process. This shift significantly enhances the overall execution efficiency, as the agent can bypass repeated reasoning processes and rely on pre-determined, efficient action sequences.
\section{Experiments}
In this section, we will present our evaluation of the evolutionary framework through a combination of various experiments on multiple benchmarks.

\subsection{Experimental Setup}

\noindent\textbf{Evaluation metrics.}
To ensure a fair and accurate comparison of our proposed method with baseline models and existing works, we adopt several evaluation metrics, which have been commonly used in prior research \cite{AppAgent, MobileAgent, appagentv2}. The metrics we report and compare are as follows:

\noindent\ding{226}\textbf{Average Steps per Task (Steps):} This measures the average number of operations the agent performs to complete a task.

\noindent\ding{226}\textbf{Average Overall Success Rate (SR):} This metric evaluates the proportion of tasks completed successfully by the agent.

\noindent\ding{226}\textbf{Average Task Time (Task Time):} The total time taken to complete a task, starting from the initiation of the task execution process to the determination of task completion by the agent.

\noindent\ding{226}\textbf{Average Step Time (Step Time):} The average time consumed per operation (step) during task execution.

\noindent\ding{226}\textbf{Average LLM Token Consumption (Tokens):} The total number of tokens used by the language model (LLM) during the task execution, including both prompt tokens and completion tokens.

For time-related comparisons, we focus only on tasks that are successfully executed by all methods. This ensures that the comparisons are not skewed by failures due to early terminations or excessive retry attempts, which might distort the results. To calculate token consumption, we compute the total number of prompt and completion tokens used by the LLM for each task and report the average number across all tasks.
To mitigate the impact of random fluctuations, each task is repeated five times by default, and we report the averaged results across these repetitions. This helps ensure that our findings are statistically robust and not affected by outlier performance from individual task executions.

\begin{table*}[t]
    \caption{\textbf{Analysis of Different Components in AppAgentX.} This table compares the performance differences resulting from the different designs with the baseline. In that table for the GPT-4o approach, we use direct LLM invocation. Both our memory design and evolution mechanism can improve success rate and efficiency.}
    \label{tab:overall_perfor}
    \small
    \centering
    \begin{tabular}{c c c c m{1.5cm}<{\centering} m{1.2cm}<{\centering} c}
    \toprule 
    \textbf{Method} & \textbf{Memory Type} & \textbf{Action Space} & \textbf{Steps\(\downarrow\)} & \textbf{Step Time (s)\(\downarrow\)} & \textbf{Tokens (k)\(\downarrow\)} & \textbf{SR \(\uparrow\)} \\
    \midrule 
    \\[-1.0em]
    GPT-4o     & None     & Basic          & 10.8 & 26 & 6.72 & 16.9\% \\
    AppAgent   & Element  & Basic          & 9.3  & 24 & 8.46 & 69.7\% \\
    \\[-1.0em] \hline \\[-1.0em]
    AppAgentX  & Chain    & Basic          & 9.1  & 23 & 9.26 & 70.8\% \\
    AppAgentX  & Chain    & Basic+Evolve   & \textbf{5.7}  & \textbf{16} & \textbf{4.94} & \textbf{71.4\%} \\
    \bottomrule
\end{tabular}
\end{table*}

\noindent\textbf{Benchmarks.}
We evaluate the performance of our method on several widely used benchmarks to validate its effectiveness and generalizability. These benchmarks cover a diverse range of tasks and applications, providing a comprehensive assessment of our framework’s capabilities. The benchmarks used in our experiments include:

\noindent\ding{226}\textbf{AppAgent Benchmark \cite{AppAgent}:} This benchmark consists of 50 tasks across 9 different applications, including 45 general tasks and 5 long-duration tasks.

\noindent\ding{226}\textbf{DroidTask \cite{DroidTask}:} Comprising 158 high-level tasks derived from 13 widely used mobile applications.

\noindent\ding{226}\textbf{AndroidWorld \cite{AndroidWorld}:} AndroidWorld is a fully functional and dynamic Android benchmark that supports 116 programmatic tasks across 20 widely used real-world Android applications. 

\noindent\ding{226}\textbf{A3 (Android Agent Arena) \cite{Athree}:} It consists of 201 tasks derived from 20 widely used third-party applications, covering common user scenarios. The benchmark provides a comprehensive evaluation mechanism, which significantly accelerates our experimental work.
\noindent\textbf{Implementation details.}
The implementation of our framework leverages several key platforms. For the foundational LLM, we selected GPT-4o \cite{GPT-4} as the default model unless otherwise stated. The LangGraph framework \cite{LangGraph} is used as the agent platform, providing essential features for LLM input-output parsing and process control. To implement the memory mechanism, we integrated Neo4j \cite{Neo4j} for graph-based storage and retrieval and Pinecone \cite{Pinecone} for vector search. For feature matching, we employed cosine similarity with embeddings derived from ResNet-50 \cite{he2015deepresiduallearningimage}, which enables effective task representation and retrieval. All experiments were conducted using the Android Debug Bridge (ADB), enabling on-device evaluations for mobile applications.

\subsection{Experimental Analysis}
In this part, the experiments are designed to validate AppAgentX's advantages in terms of efficiency and accuracy.
We conducted five experiments for the baseline comparison and two experiments for the large dataset to mitigate the effects of randomness in LLMs.

\noindent\textbf{Comparative Analysis.}
A comparison of our experimental results with model variants is shown in Table \ref{tab:overall_perfor}. We report the results on the AppAgent benchmark.
We begin with a baseline model that does not incorporate any memory mechanism and progressively introduce our proposed enhancements. The first model variant integrates a module that records only information about elements, providing a basic form of memory. Subsequently, we introduce a more advanced memory design that maintains a structured memory chain, enabling the model to retain a more comprehensive representation of past interactions. Finally, we augment the system with an evolutionary mechanism that expands the action space to include high-level actions, further optimizing task execution.

As the result shows, incorporating the memory mechanism significantly improves the task success rate. The baseline model, which lacks memory, achieves an SR of only 16.9\%. 
Introducing the chain-structured memory significantly enhances the success rate, reaching 70.8\%, highlighting the clear advantage of maintaining a more structured history of interactions. Furthermore, it proves to be more effective than element memory in facilitating task completion.
Moreover, the integration of the evolutionary mechanism leads to substantial efficiency gains. Expanding the action space to include high-level actions decreases the average required number of steps from 9.1 to 5.7, while the step execution time is reduced from 23 to 16 seconds. Additionally, average token consumption is significantly minimized, dropping from 9.26k to 4.94k, indicating a more efficient decision-making process. Notably, this enhancement reduces computational overhead and improves the average success rate to 71.4\%.
These results show that our approach is effective. The memory mechanisms significantly improve task performance, while the evolutionary strategy boosts efficiency by reducing steps, execution time, and token use.

\begin{table*}[t]
    \caption{\textbf{Comparison with AppAgent on Large Benchmarks.} This table evaluates the efficiency and accuracy of different frameworks on benchmarks containing a large number of tasks. }
    \label{tab:benchmark_comparison}
    \centering
    \small
    \begin{tabular}{lclccc}
    \toprule
    \\[-1.0em]
    \textbf{Benchmarks} & \textbf{Task Num.} & \textbf{Framework} & \textbf{Task Time\(\downarrow\)} & \textbf{Tokens (k)\(\downarrow\)} & \textbf{SR\(\uparrow\)} \\
    \\[-1.0em]
    \midrule
    \\[-1.0em]
    \multirow{2}{*}{DroidTask\cite{DroidTask}} & \multirow{2}{*}{158} & AppAgent       & 106.24 & 11.5 & 46.3\% \\
                               &                      & AppAgentX      & \textbf{56.29} & \textbf{5.1} & \textbf{88.2}\% \\
    \\[-1.0em] \hline \\[-1.0em]
    \multirow{2}{*}{AndroidWorld\cite{AndroidWorld}} & \multirow{2}{*}{116} & AppAgent       & 147.17 & 18.9 & 41.7\% \\
                               &                      & AppAgentX      & \textbf{59.74}  & \textbf{6.2} & \textbf{62.5}\% \\
    \\[-1.0em] \hline \\[-1.0em]
    \multirow{2}{*}{A3 (Android Agent Arena)\cite{Athree}} & \multirow{2}{*}{201} & AppAgent       & 134.67 & 19.2 & 10.3\% \\
                               &                      & AppAgentX      & \textbf{48.12}  & \textbf{4.7} & \textbf{39.3}\% \\
    \\[-1.0em]
    \bottomrule
\end{tabular}
\end{table*}
To further highlight the advantages of the AppAgentX framework, we compare its performance to other state-of-the-art frameworks on additional benchmarks, including DroidTask \cite{DroidTask}, Android Agent Arena\cite{Athree} and AndroidWorld\cite{AndroidWorld}. As shown in Figure \ref{tab:benchmark_comparison}, AppAgentX outperforms AppAgent, in both task execution time and accuracy. Specifically, AppAgentX achieves significantly higher efficiency while maintaining higher task success rates across a broader range of applications.
\begin{figure}[t]
    \centering
    \includegraphics[width=\linewidth]{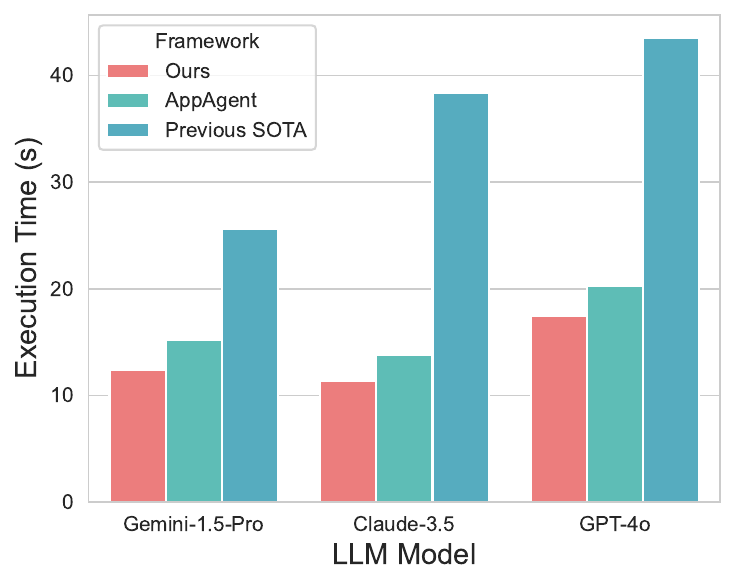}
    \caption{\textbf{Comparison of Average Execution Time per Step.} This figure presents the average execution time per steps across different LLMs and frameworks.}
    \label{fig:llm_comparison}
\end{figure}

Additionally, we compare the execution efficiency of AppAgentX against two other frameworks \cite{mobileagent2,AppAgent} across several prominent foundational LLMs, including GPT-4o \cite{GPT-4}, Claude 3.5 Sonnet \cite{Claude}, and Gemini 1.5 Pro \cite{Gemini1.5}. While acknowledging that certain discrepancies in the experimental setup may exist, we focus our evaluation primarily on execution time, as it serves as a practical and reproducible metric for assessing efficiency. As shown in Table \ref{fig:llm_comparison}, AppAgentX consistently demonstrates faster per-step completion times compared to the previous state-of-the-art \cite{mobileagent2} and AppAgent, across multiple LLM backends. These results suggest improved efficiency and better alignment with real-world deployment constraints.

\noindent\textbf{Task difficulties analysis.} 
\begin{figure}[t]
    \centering
    \includegraphics[width=0.8\linewidth]{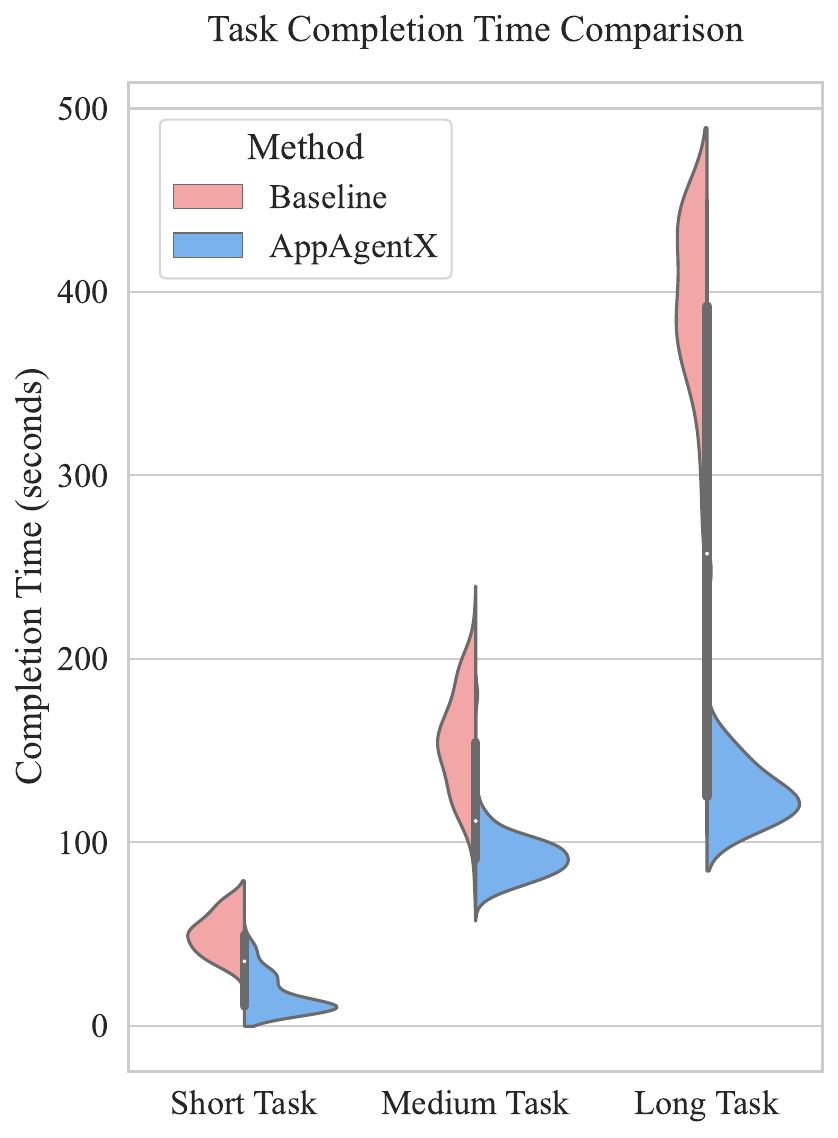}
    \vspace{-1em}
    \caption{\textbf{Task Completion Times Across Different Task Lengths.} This figure shows the distribution of task completion times for short, medium, and long tasks. Each violin plot represents the density of completion times, with wider sections indicating higher data concentration. AppAgentX consistently outperforms the baseline, particularly for longer tasks.}
    \vspace{-0.5em}
    \label{fig:length_test_result}
\end{figure}
\begin{figure*}[t]
    \centering
    \includegraphics[width=0.9\linewidth]{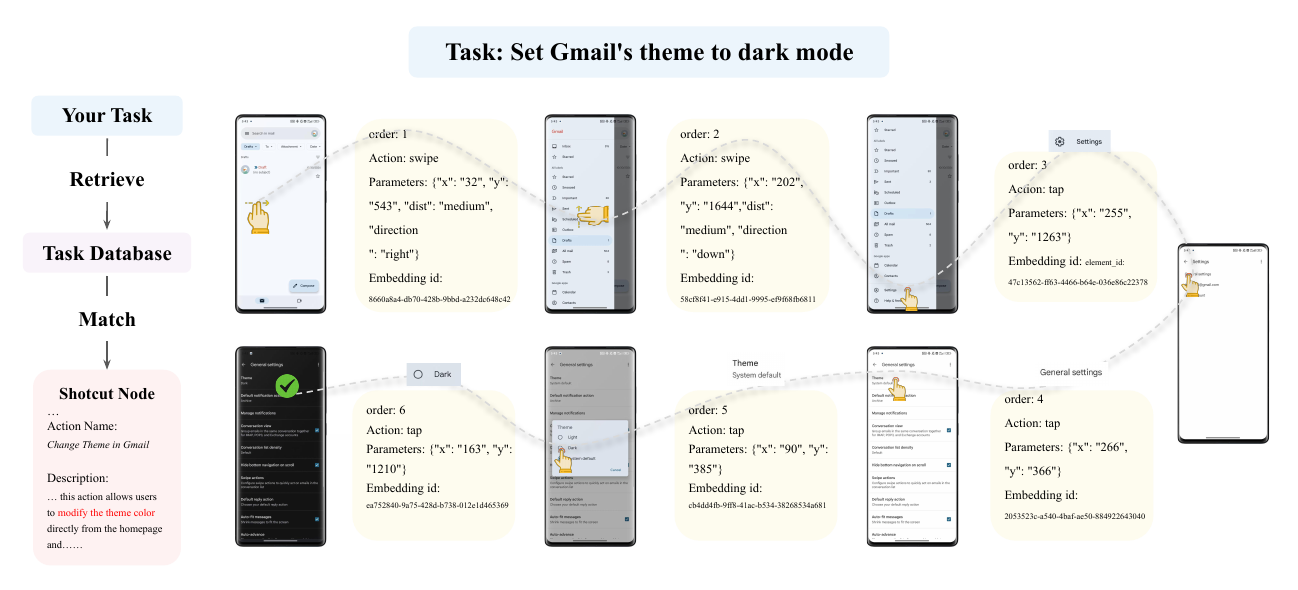}
    \caption{\textbf{Qualitative Results.} This figure presents the path execution utilizing shortcuts on Gmail. It demonstrates the efficiency of AppAgentX in decreasing the utilization of LLM for per-step reasoning.}
    \label{fig:combined}
\end{figure*}
To investigate the performance of AppAgentX across tasks of varying complexity, we categorize the tasks based on their ground-truth operation steps, annotated through expert demonstrations. Tasks with up to 5 steps are considered short, tasks with 6 to 10 steps are classified as medium-length, and tasks requiring more than 10 steps are categorized as long tasks.
As shown in Figure \ref{fig:length_test_result}, the violin plot provides an intuitive visualization of how task complexity affects performance. Notably, our method demonstrates a clear advantage in terms of task execution time as the task complexity increases. To account for random fluctuations, we conducted one-tailed paired t-tests \cite{ttest} on each data group to verify that AppAgentX significantly outperforms the baseline in efficiency, with all p-values falling below 0.05. Our model consistently outperforms the baseline across all task lengths, resulting in reduced task completion times. This finding reinforces the robustness and efficiency of our approach, confirming its stability under varying task difficulties.

\begin{table}[t]
    \caption{\textbf{Ablation Study of Different Screen Perceptrons and Action Spaces.} 
    This table compares task success rates and times for different screen parsers and two action space variants: 
    \textbf{BAS} (Basic Action Space) and \textbf{FAS} (Full Action Space).}
    \centering
    \small
    \begin{tabular}{lcc} 
        \toprule
        \textbf{Methods} & \textbf{Task Time\(\downarrow\)} & \textbf{SR\(\uparrow\)} \\ 
        \midrule
        Omniparser + BAS         & 209.3s & 70.8\% \\
        Omniparser + FAS         & 91.2s  & 71.4\% \\
        Ferret-UI + BAS          & 201.7s & 68.1\% \\
        Ferret-UI + FAS          & 92.5s  & 70.6\% \\
        \bottomrule
    \end{tabular}
    \label{tab:omni_ablation}
\end{table}

\noindent\textbf{Ablation Study.}
To further evaluate the performance gains of our agent, we substitute different screen perceptrons to assess the effectiveness of our evolutionary approach. Specifically, OmniParser\cite{OmniParser} is employed as the default visual element extractor to locate UI components on the screen. In this study, we replace it with Ferret-UI\cite{FerretUI}, a component with equivalent functionality, and compare the baseline action space with the evolved action space to examine their respective impacts.

As shown in Table \ref{tab:omni_ablation}, the Full Action Space significantly improves task performance compared to the Basic Action Space, significantly reducing execution time while simultaneously improving success rates. These results validate the effectiveness of the evolved action space.

\noindent\textbf{Qualitative Analysis.}
In Figures \ref{fig:combined}, we show a qualitative analysis of some of the actions that used high-level actions to operate the screen. On the left side we label the various processes in the relevant execution operations, from the retrieval of tasks to the matching of shortcut nodes. Despite the different user interfaces and actions performed by these applications, AppAgentX successfully completed the given task.

\section{Conclusion}
We propose an evolving GUI agent framework that enhances efficiency and intelligence by abstracting high-level actions from execution history. Unlike rule-based automation, our approach generalizes task execution by compressing repetitive operations, balancing intelligent reasoning with efficient execution. A chain-based knowledge framework enables continuous behavior refinement, improving adaptability and reducing redundancy. Experiments show our framework outperforms existing methods in accuracy and efficiency.

\bibliography{custom}

\appendix

\section{Design and Execution Details}
This section presents the detailed design of the proposed chain structure and the overall task execution workflow in our system.
\subsection{Chain Design Details}

In chain design, there are three types of relationships between nodes:

\noindent \textbf{Page-HAS\_ELEMENT$\rightarrow$Element}

\textbf{Type}: HAS\_ELEMENT

\textbf{Direction}: (Page) -[:HAS\_ELEMENT]-> (Element)

\textbf{Purpose}: Indicates that a page contains a specific UI element.

\noindent \textbf{Shortcut-COMPOSED\_OF$\rightarrow$Element}

\textbf{Type}: COMPOSED\_OF

\textbf{Direction}: (Shortcut) -[:COMPOSED\_OF]-> (Element)

\textbf{Relationship Attributes}:
\begin{itemize}
    \item \textbf{order}: Integer, representing the execution sequence.
    \item \textbf{atomic\_action}: Type of basic action (e.g., click, text input, etc.).
    \item \textbf{action\_params}: JSON format, potentially including input text, click parameters, etc.
\end{itemize}

\textbf{Purpose}: Defines how a composite action consists of certain elements and their corresponding steps, which must be executed in a specific order.

\noindent \textbf{Element-LEADS\_TO$\rightarrow$Page}  

\textbf{Type}: LEADS\_TO

\textbf{Direction}: (Element) -[:LEADS\_TO]-> (Page)
\textbf{Purpose}: Represents the linkage of a composite action.

Together, these types of edges, along with the nodes described in the main text, constitute the structure of the complete memory store. This modeling approach aligns with the relationships between page jumps and containment, and offers a solid foundation for analyzing the evolution of actions.

\subsection{Execution Process Details}
\begin{figure}[t]
    \centering
    \includegraphics[width=1\linewidth]{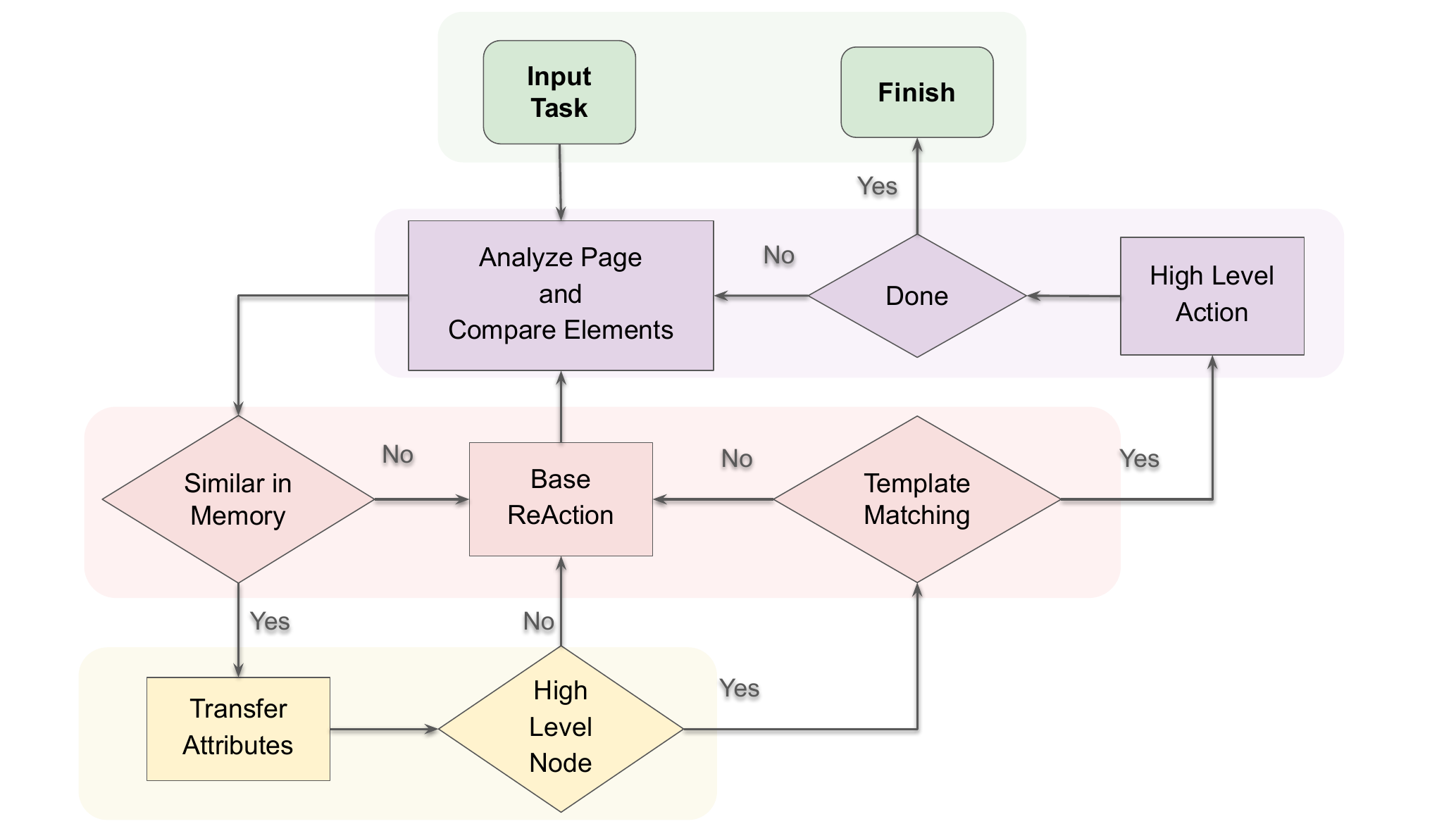}
    \caption{\textbf{Overall Execution Process.} This figure illustrates the flowchart from the input of the task to the final execution result after we then add the shortcut route.}
    \label{fig:flow}
\end{figure}
Figure \ref{fig:flow} presents a detailed flowchart that delineates the task processing workflow. The primary components of this workflow include page analysis, element comparison, matching, and memory storage.
As depicted in Figure \ref{fig:flow}, the system utilizes a hierarchical action framework comprising high-level nodes and basic action spaces. During the operation, if the matching retrieval process of a high-level node fails, the system seamlessly transitions to the corresponding basic action spaces. 

\section{Detailed Numerical Results }
This section provides the numerical analysis results of various shortcut designs, including evaluations of robustness and statistical tests to validate the effectiveness of the proposed system design.

To supplement the grouped bar chart presented in Figure~\ref{fig:llm_comparison}, this subsection provides the exact per-step execution time (in seconds) for each evaluated framework (MobileAgent2, AppAgent, and AppAgentX) across different LLM backends. These numerical results are intended to enhance reproducibility, facilitate further analysis, and offer a precise basis for comparing execution efficiency across systems.

\begin{table}[h]
    \caption{\textbf{Comparison of Average Execution Time per Step.} This table presents the average execution time (seconds \(\downarrow\)) across different LLMs and frameworks.}
    \label{tab:llm_comparison}
    \centering
    \small
    \begin{tabular}{llll} 
        \toprule
        \textbf{LLM} & \textbf{Previous SOTA} & \textbf{AppAgent} & \textbf{Ours} \\
        \midrule
        Gemini-1.5-Pro & 25.6 & 15.2 & \textbf{12.4} \\
        Claude-3.5     & 38.4 & 13.8 & \textbf{11.4} \\
        GPT-4o         & 43.5 & 20.3 & \textbf{17.5} \\
        \bottomrule
    \end{tabular}
\end{table}

Table~\ref{tab:llm_comparison} lists the detailed values corresponding to each bar in the figure.

\begin{table}[h]
    \caption{\textbf{Statistical Test Results for Task Completion Times.} Results from one-tailed paired t-tests.}
    \label{tab:statistical_tests}
    \centering
    \small
    \begin{tabular}{lccc} 
        \toprule
        \textbf{Task Length} & \textbf{Mean Difference (s)} & \textbf{t-value} & \textbf{p-value} \\ 
        \midrule
        Short Task  & -12.4  & -3.45 & <0.01 \\
        Medium Task & -30.7  & -5.12 & <0.001 \\
        Long Task   & -75.3  & -6.89 & <0.001 \\
        \bottomrule
    \end{tabular}
\end{table}

\begin{figure*}[t]
    \centering
    \includegraphics[width=0.9\linewidth]{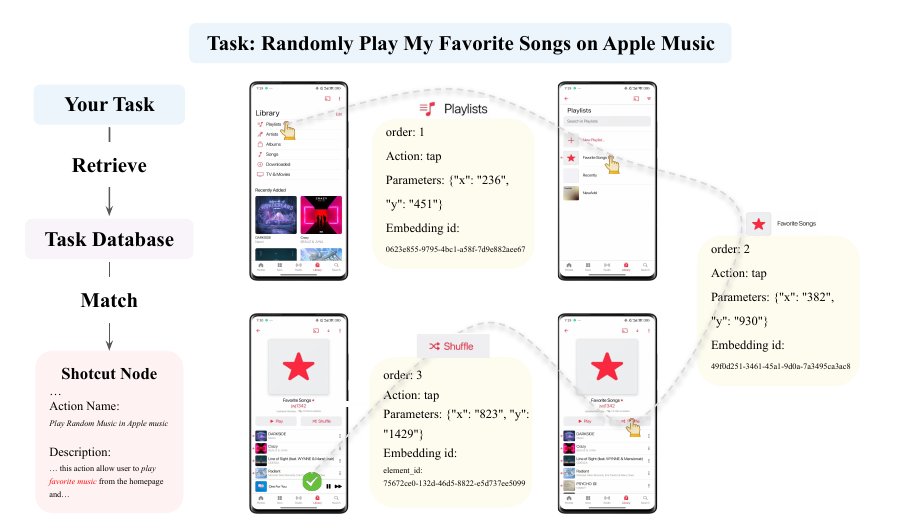}
    \caption{\textbf{Qualitative Results.} This figure presents the path execution utilizing shortcuts on Apple Music. It demonstrates the efficiency of AppAgentX in decreasing the utilization of LLM for per-step reasoning.}
    \label{fig:app_combined}
\end{figure*}

To reinforce the observed differences in task completion times, we complemented the analysis with detailed results from one-tailed t-tests, as summarized in Table~\ref{tab:statistical_tests}. These results address data not fully elaborated upon in the main text. The negative t-values reported in this table indicate the presence of a directional effect, where AppAgentX consistently outperforms the baseline in reducing task completion times. Specifically, the negative sign reflects that the mean completion time for AppAgentX is significantly lower than that of the baseline across all task lengths. This directional result aligns with our hypothesis and further validates the consistent efficiency advantage of AppAgentX.
\section{Additional Quantitative Analysis}
To further illustrate the behavior of our system in real-world settings, we include a qualitative example in Figure~\ref{fig:app_combined}. This example showcases the execution path for a task in Apple Music, highlighting how AppAgentX leverages existing shortcuts to complete the task more efficiently. By using pre-defined shortcuts, the system reduces the reliance on large language model (LLM) reasoning at each step, thereby improving both speed and resource usage.

\end{document}